\documentclass[11pt]{article}

\usepackage{dialogue}
\usepackage[russian,british]{babel}

\usepackage{ifxetex}
\ifxetex
    \usepackage{fontspec}
    \setromanfont{Times New Roman}
\else
  \usepackage{cmap}
  \usepackage[T2A,T1]{fontenc}
  \usepackage[utf8]{inputenc}
  \usepackage{times}
  \usepackage{latexsym}
  \usepackage{substitutefont}
  \substitutefont{T2A}{\familydefault}{cmr}
\fi

\usepackage{graphicx}
\usepackage{tabularx}
\usepackage{multirow}

\usepackage{url}
\usepackage{pgf}

\usepackage{covington} 

\usepackage{hyperref}

\dialogfinalcopy 

\title{Argumentative Text Generation \\ in Economic Domain}

\author{Fishcheva I. N. \\
 Vyatka State University \\
 Kirov, Russia \\
 {\tt fishchevain@gmail.com} \\\And
 Osadchiy D. \\
 ITMO University \\
 Saint Petersburg, Russia \\
 {\tt da.osadchiy@gmail.com}  \\\AND
 Bochenina K. O. \\
 ITMO University \\ 
 Saint Petersburg, Russia \\
 {\tt kbochenina@itmo.ru}\And
 Kotelnikov E. V. \\
 Vyatka State University \\
 Kirov, Russia \\
 {\tt kotelnikov.ev@gmail.com}\\}

\date{}

\begin{document}
\maketitle
\begin{abstract}
  The development of large and super-large language models, such as GPT-3, T5, Switch Transformer, ERNIE, etc., has significantly improved the performance of text generation. One of the important research directions in this area is the generation of texts with arguments. The solution of this problem can be used in business meetings, political debates, dialogue systems, for preparation of student essays. One of the main domains for these applications is the economic sphere.\par
		The key problem of the argument text generation for the Russian language is the lack of annotated argumentation corpora. In this paper, we use translated versions of the Argumentative Microtext, Persuasive Essays and UKP Sentential corpora to fine-tune RuBERT model. Further, this model is used to annotate the corpus of economic news by argumentation. Then the annotated corpus is employed to fine-tune the ruGPT-3 model, which generates argument texts. The results show that this approach improves the accuracy of the argument generation by more than 20 percentage points (63.2\% vs. 42.5\%) compared to the original ruGPT-3 model.
  
  
  \textbf{DOI:} 10.28995/2075-7182-2022-21-211-222
\end{abstract}

\selectlanguage{russian}
\begin{center}
  \russiantitle{Генерация аргументативных текстов экономической тематики}

  \medskip \setlength\tabcolsep{0.5cm}
  \begin{tabular}{cc}
    \textbf{Фищева И. Н.} & \textbf{Осадчий Д.}\\
      Вятский государственный университет & Университет ИТМО\\
      Киров, Россия & Санкт-Петербург, Россия \\
      {\tt fishchevain@gmail.com} &  {\tt da.osadchiy@gmail.com} \\ \\
      \textbf{Боченина К. О.} & \textbf{Котельников Е. В.}\\
      Университет ИТМО & Вятский государственный университет\\
      Санкт-Петербург, Россия & Киров, Россия \\
      {\tt kbochenina@itmo.ru} &  {\tt kotelnikov.ev@gmail.com} \\
  \end{tabular}
  \medskip
\end{center}

\begin{abstract}
  Разработка больших и сверхбольших языковых моделей, таких как GPT-3, T5, Switch Transformer, ERNIE и др., позволила в последнее время значительно повысить качество генерации текстов. Одним из важных направлений в этой области является порождение текста с аргументами. Решение такой задачи может быть использовано при проведении деловых совещаний, в политических дебатах, в диалоговых системах, при подготовке студенческих эссе. Одной из основных предметных областей в указанных приложениях является экономическая сфера.
  
  Ключевой проблемой при генерации аргументов для русского языка является дефицит корпусов, размеченных по аргументации. В настоящей работе мы используем переводные версии корпусов Argumentative Microtext, Persuasive Essays и UKP Sentential для дообучения модели RuBERT. Далее построенная модель используется для разметки по аргументации собранного корпуса экономических новостей. Затем размеченный корпус применяется для дообучения модели ruGPT-3, которая порождает аргументационные тексты. Результаты показывают, что такой подход позволяет повысить правильность генерации аргументов более чем на 20 процентных пунктов (63.2\% vs. 42.5\%) по сравнению с исходной моделью ruGPT-3.

  \textbf{Ключевые слова:} анализ аргументации, генерация текстов, BERT, GPT
\end{abstract}
\selectlanguage{british}

\section{Introduction}
	Automatic text generation has recently made impressive progress with the development of large and super-large pre-trained language models \cite{Han2021}, such as GPT-3 \cite{Brown2020}, T5 \cite{Raffel2020}, Switch Transformer \cite{Fedus2021}, ERNIE \cite{Sun2021} and others. These models allow tuning to the problem by updating the weights on a small training dataset (fine-tuning) or without updating weights in few-shot learning (with several training examples, usually from 10 to 100), one-shot learning (one training example) and even zero-shot learning \cite{Brown2020}. 
	
	One of the important directions in this field is the generation of texts with arguments \cite{Hua2020,Gretz2020,Schiller2021}. Arguments in this case are either found using an information retrieval system \cite{Hua2020}, or generated by a pre-trained language model \cite{Gretz2020,AlKhatib2021}. \textit{An argument} is a set of statements that includes claim and premises \cite{Stede2018}. \textit{A claim} is a statement that expresses a potentially controversial point of view. \textit{Premises} are statements supporting or refuting this claim.
	
	Systems that generate text with arguments related to a given claim can be used during business meetings to promptly generate arguments; in political debate to back up the position of a speaker; in jurisprudence to search and generate arguments for legislative acts and precedents; in dialogue systems for the selection of arguments to convince the interlocutor; in education for analyzing, generating, and evaluating arguments in student essays. One of the most common domains for these applications is economics. 
	
	There are several publicly available pre-trained language models for the Russian language, including RuBERT \cite{Kuratov2019}, SBERT \cite{SBERT}, and ruGPT-3 \cite{ruGPT-3}. Recently, several papers have appeared on the argument mining in Russian \cite{Fishcheva2019,Fishcheva2021,Salomatina2021,Ilina2021}.
	
	However, to the best of our knowledge, there have been no work devoted to the generation of argumentative Russian texts so far. We aim to close this gap and propose an approach to build a language model for generating an argumentative text in Russian in the economic domain.
	
	The contribution of this work is as follows:
	\begin{itemize}
		\item a new approach to the generation of Russian argumentative texts is proposed. In this approach, the model RuBERT is first trained using argumentative corpora, which is then used to automatically annotate arguments in an economic text corpus. The annotated corpus is used to fine-tune the ruGPT-3 model, which allows generating texts with arguments, particularly, sentences containing premises related to a given claim;
		\item presenting the results of experiments with the proposed approach;
		\item sharing the corpus of pairs <\textit{claim, sentence}> on economic domain. The corpus includes 660 sentences manually annotated into two classes – “premise” / “non-premise”.
	\end{itemize}

\section{Previous Work}
	\subsection{Argumentative Text Generation}
	Argumentative texts generation includes the following research directions.
	\begin{enumerate}
		\item Application of traditional text generation architectures, including content selection/organization and linguistic implementation components \cite{Carenini2006}.
		\item Generation of an argumentative text from given small argumentative elements \cite{WaltonD.2012,Reisert2015,Wachsmuth2018,ElBaff2019}.
		\item Replacing target objects in extracted arguments \cite{Bilu2016}.
		\item Using predefined argument templates \cite{Bilu2020}.
		\item Application of the encoder-decoder architecture (seq2seq) \cite{Hidey2019}, possibly supplemented by an information retrieval component \cite{Hua2020,Schiller2021}.
		\item Using pre-trained language models for text generation, such as GPT-2 \cite{Gretz2020,AlKhatib2021}.
	\end{enumerate}
	The papers closest to our research are \cite{Schiller2021,Gretz2020,AlKhatib2021}.
	
	Schiller et al. \cite{Schiller2021} proposed a neural network model Arg-CTRL for controlled generation of texts based on the well-known CTRL model \cite{Keskar2019}. Controlling the text generation is carried out using the so-called \textit{control codes}, including the topic, stance, and aspect of the argument. The retrieval system and the BERT model were used to extract aspects. In our work, instead of the CTRL model, we use the ruGPT-3 model, which does not require learning from scratch for the Russian language, like CTRL. Also, the BERT model is used not to extract aspects, but to select sentences containing premises.
	
	Gretz et al. \cite{Gretz2020} investigated claim generation. They suggested using the fine-tuning of the neural network model GPT-2 \cite{Radford2020} based on the corpus of automatically extracted arguments. Moreover, authors added context to the input data for GPT-2 in order to improve the quality of generation. In our work, we generated premises, not claims. Also, instead of GPT-2, which is mainly an English-language model, we use the Russian-language ruGPT-3 model.
	
	Al-Khatib et al. \cite{AlKhatib2021} used argumentation knowledge graphs to generate argumentative texts. The concepts contained in such graphs are searched on debate portals and Wikipedia. The found fragments of texts are used for fine-tuning GPT-2. In our work, we use the RuBERT model rather than knowledge graphs to generate training data. In addition, instead of GPT-2 we use the Russian-language model ruGPT-3.
	
	\subsection{Argumentation Mining in Russian}
	The area of argumentation mining for the Russian language has recently attracted more and more attention of researchers.
	
	The first annotated corpus for the Russian language \cite{Fishcheva2019} was created based on the translation of the English language Argumentative Microtext Corpus (ArgMicro) \cite{Peldszus2015,Skeppstedt2018}. It was then expanded with machine translation of the Persuasive Essays Corpus (PersEssays) \cite{Stab2014} and a Joint Argument Annotation Scheme was proposed \cite{Fishcheva2021}. By using XGBoost and BERT, the authors were able to improve the results of automatic classification of “for” / “against” premises.
	
	Salomatina et al. \cite{Salomatina2021} proposed a method for finding an argumentative structure based on using the patterns of argumentation indicators and their role in the thematic structure of the text. This method can be used in the absence of a sufficient amount of annotated data.
	
	Ilina et al. \cite{Ilina2021} presented a web resource designed to study argumentation in popular science discourse. A distinctive feature of the argument annotation model is the using of argument persuasiveness weighting. The annotation script includes several procedures that allow the annotator to check the quality of text annotation and evaluate the persuasiveness of the argumentation.
	
	In our work, to the best of our knowledge, the problem of generating argumentative texts in Russian is being investigated for the first time. We are expanding the Russian-language corpus from \cite{Fishcheva2021} by translating the UKP Sentential Argument Mining Corpus (UKP Sentential) \cite{Stab2018}. Based on the extended corpus, we train RuBERT model, which we then use to annotate sentences of the economic news corpus. The annotated corpus is used to fine-tune the ruGPT-3 model to generate texts containing premises for a given claim.

\section{Proposed approach}
\begin{figure}
	\begin{center}
		\includegraphics[width=1\textwidth]{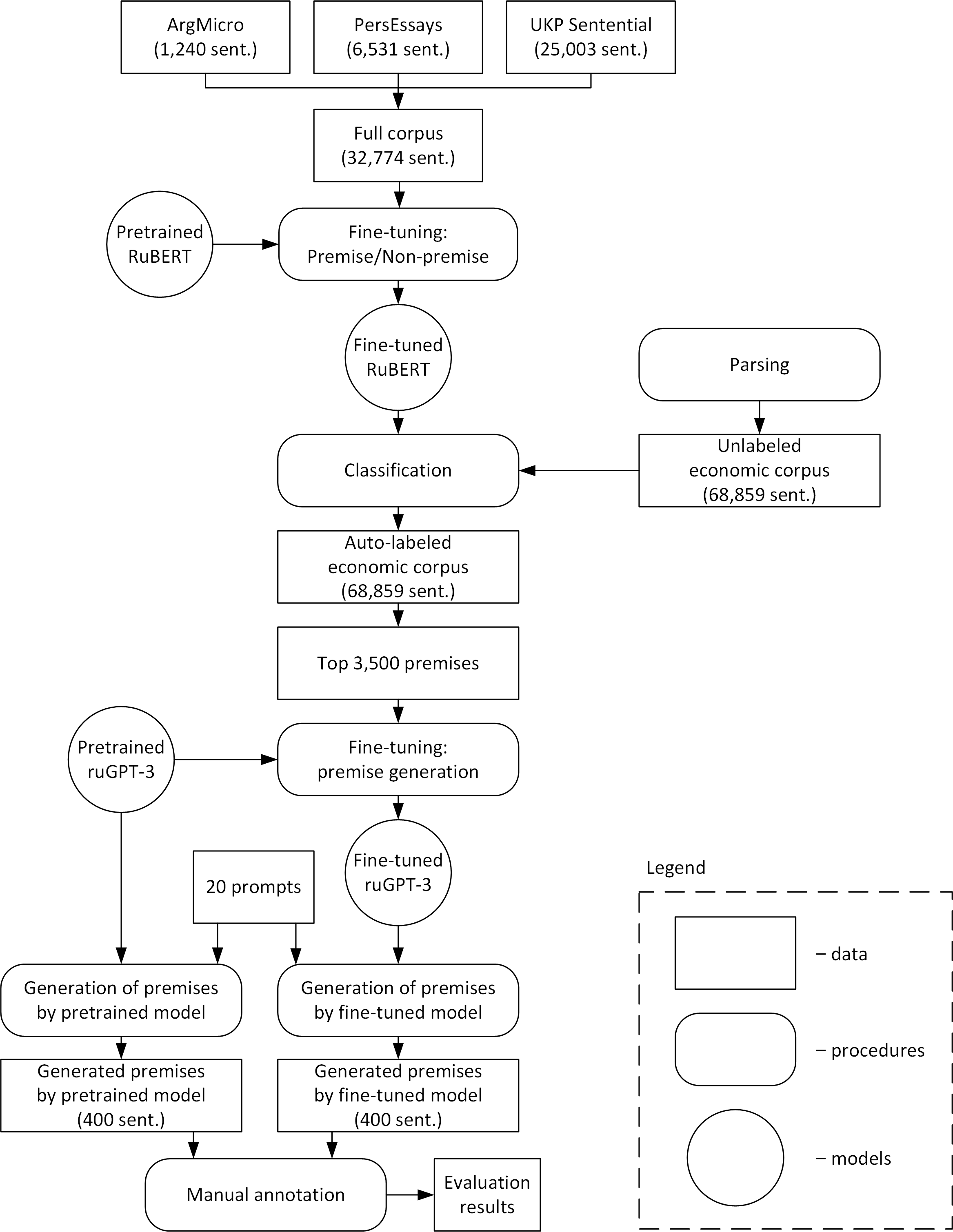}
		\caption{Scheme of the proposed approach.} \label{fig1}
	\end{center}
	\end{figure}

The scheme of the proposed approach for generating argumentative texts is shown in Fig.~\ref{fig1}.
	
	The three argumentative annotated corpora – ArgMicro, PersEssays, and UKP Sentential (see Section~\ref{text_corpora}) – serve as input data. They are used to fine-tune the pre-trained RuBERT\footnote{\href{https://huggingface.co/DeepPavlov/rubert-base-cased}{https://huggingface.co/DeepPavlov/rubert-base-cased}} model \cite{Kuratov2019}. This model showed the best results (along with XGBoost\footnote{\href{https://xgboost.readthedocs.io}{https://xgboost.readthedocs.io}} model) in the argument classification task \cite{Fishcheva2021}. We also compared the performance of RuBERT with XGBoost on “premise” / “non-premise” classification task at the sentence level. As a result, RuBERT outperformed XGBoost (see Subsection~\ref{classification}) and further we used the RuBERT model.
	
	We collected the corpus of economic news from the internet (see Section~\ref{text_corpora}) -- it is denoted as “Unlabeled economic corpus” on the scheme. The fine-tuned RuBERT classifies economic corpus sentences so we obtain economic corpus labeled by “premise” / “non-premise” (see details in Subsection~\ref{classification}).
	
	At the next step, fine-tuning of the pretrained ruGPT-3 \cite{ruGPT-3} model is carried out. Firstly, we choose 3,500 sentences (it is about 5\% from the whole corpus) with the highest probability scores of classification as “premise” from RuBERT. Then we take pretrained ruGPT-3 model and fine-tune it on selected 3,500 sentences (see details in Subsection~\ref{rugpt-3}). To evaluate and compare the accuracy of the original model and the fine-tuned one, 20 prompts were used, for each prompt both models generated 20 potential premises. The 800 (400+400) received sentences were annotated manually (see details in Subsection~\ref{manual_annotation}).

\section{Text Corpora} \label{text_corpora}
	In our study, we used three existing argument-annotated corpora – ArgMicro, PersEssays, and UKP Sentential, as well as a new economic news corpus.
	
	The Argumentative Microtext Corpus (ArgMicro) was proposed in \cite{Peldszus2015,Skeppstedt2018}. The corpus includes 283 texts on various topics (raising the retirement age, health insurance, school uniforms, etc.). Each text contains one claim about a topic and from 2 to 10 ADUs (argumentative discourse units), annotated as premises “for” or “against” this claim. An ADU is a fragment of text that has a single argumentative value \cite{Stede2018}. In the ArgMicro, an ADU can be either an entire sentence or part of a sentence. Further, we consider each individual ADU as a sentence.
	
	The Persuasive Essays Corpus (PersEssays) was introduced in \cite{Stab2014}. It contains 399 texts on a wide range of topics (school education, immigration, economic policy, etc.). The ADUs are sentences. The annotation of sentences is done according to four types: main claim, claim, premise, and neutral element. When forming the training corpus for the classification “premise” / “non-premise”, the main claims were excluded, just claims and the actual premises were used as “premises”, and neutral elements – as “non-premises”.
	
	The UKP Sentential Argument Mining Corpus (UKP Sentential) was proposed in \cite{Stab2018}. The corpus includes 25,492 sentences, annotated as “for”, “against” or “non-premise” in relation to one of the eight topics (abortion, cloning, the death penalty, etc.). The corpus contains 489 sentences, the annotation of which is different for different topics. Such sentences were excluded.
	
	\cite{Fishcheva2019} showed that the best result among the Google Translate, Yandex.Translate and Promt systems was demonstrated by Google Translate during machine translation of the ArgMicro corpus from English into Russian. Therefore, all three English corpora were translated into Russian using Google Translate.
	
	The characteristics of all the corpora with argumentative annotation are shown in Table~\ref{tab_corpora}.
	\begin{table}
		\begin{center}
			\caption{Characteristics of corpora with argumentative annotation.}
			\renewcommand\tabcolsep{5.0pt}
			\label{tab_corpora}
			\begin{tabularx}{\textwidth}{| X | X | X | X |}
				\hline
				Corpora & Premise & Non-premise & Total\\
				\hline
				ArgMicro & 1,236 & 4 & 1,240 \\ 
				PersEssays & 5,302 & 1,229 & 6,531 \\ 
				UKP Sentential & 11,023 & 13,980 & 25,003 \\ 
				\hline
				Total & 17,561 & 15,213 & 32,774 \\
				\hline
			\end{tabularx}
		\end{center}
	\end{table}
	
	For fine-tuning of the ruGPT-3 model in economics, we used the corpus of Russian-language economic news, collected from the website \textit{Banki.ru}\footnote{\href{https://www.banki.ru}{https://www.banki.ru}}. The original corpus included 7,759 texts for the period from 01.06.2019 to 12.07.2021. The texts were segmented into sentences using the \textit{Stanza}\footnote{\href{https://stanfordnlp.github.io/stanza}{https://stanfordnlp.github.io/stanza}} library (average text size – 9 sentences) and tokenized using the \textit{NLTK}\footnote{\href{https://www.nltk.org}{https://www.nltk.org}} library. Further, 10\% of the longest and shortest sentences, as well as repetitive sentences were removed. The preprocessing resulted in 68,859 sentences with an average length of 15 words.

\section{Result and Discussion} \label{results}
\subsection{“Premise” / “Non-premise” Classification} \label{classification}
	At the first step, we compared the performance of RuBERT and XGBoost, which were the best models in \cite{Fishcheva2021}. To this end, we fine-tuned the pre-trained RuBERT model and we trained the XGBoost classifier on the union of annotated corpora (ArgMicro $\cup$ PersEssays $\cup$ UKP Sentential). The problem of classifying sentences into two classes -- “premise” and “non-premise” -- was set. We used the same 5-fold cross-validation for both models to obtain performance scores. For selection of hyperparameters for XGBoost, 4-fold nested cross-validation was used, for RuBERT -- hold-out dataset (validation 20\%). The hyperparameters are as follows:
	\begin{itemize}
		\item XGBoost\footnote{Optimal values of the hyperparameters are highlighted in bold – they turned out to be the same on all folds.}: number of trees = [50, 150, \textbf{500}], maximum tree depth = [\textbf{2}, 8, 20, 30];
		\item RuBERT: number of epochs = [3, \textbf{5}], batch size = [4, \textbf{8}], learning rate = \textbf{$10^{-5}$}.
	\end{itemize}
    To train RuBERT, we used class \textit{BertForSequenceClassification} from the \textit{transformers}\footnote{\href{https://huggingface.co/docs/transformers/model_doc/bert}{https://huggingface.co/docs/transformers/model\_doc/bert}} library with the following settings: $AdamW$ optimizer with default parameters, weight decay = 0.01, context length = 512 tokens.
    
	Following \cite{Fishcheva2019}, we used three types of features for XGBoost:
	\begin{itemize}
		\item lexical features -- discursive markers (e.g., “therefore”, “influence”, “firstly”, etc.) and modal words (e.g., “should be”, “can”, “want”, etc.), including negations, 255 features in total;
		\item punctuation features -- comma, colon, semicolon, question and exclamation marks, 5 features in total;
		\item morphosyntactic features -- N-grams based on parts of speech (nouns, pronouns, verbs, adjectives, and adverbs), N = {2, 3, 4} (e.g., ADJ+ADJ+NOUN, ADV+VERB), and grammatical features of verbs: time, inclination, person; there are 783 features in total. Morphological analysis was carried out with \textit{Mystem}\footnote{\href{https://yandex.ru/dev/mystem}{https://yandex.ru/dev/mystem}}.
	\end{itemize}
	
	Unlike \cite{Fishcheva2019}, we have not added contextual features of the previous and next sentences, since the UKP Sentential corpus contains only individual sentences.
	
	Cross-validation results for the union corpus (ArgMicro $\cup$ PersEssays $\cup$ UKP Sentential) are presented in Table~\ref{tab_performance}.
	\begin{table}[h!]
		\begin{center}
			\caption{Performance scores for “premise” / “non-premise” classification:
				macro-averaged F1-score, Precision, Recall and Accuracy (Mean ± Std Dev)
			}\label{tab_performance}
			\renewcommand\tabcolsep{5.0pt}
			\begin{tabular}{| l | l | l | l | l |}
				\hline
				Model & F1-score & Precision & Recall & Accuracy\\
				\hline
				XGBoost & 0.6800±0.0066 & 0.6817±0.0065 & 0.6796±0.0066 & 0.6832±0.0064 \\ 
				RuBERT & \textbf{0.7903±0.0051} & \textbf{0.7901±0.0051} & \textbf{0.7908±0.0050} & \textbf{0.7911±0.0051}\\ 
				\hline
			\end{tabular}
		\end{center}
	\end{table}
	
	XGBoost is significantly inferior to RuBERT in this task, in contrast to the results of \cite{Fishcheva2021}, where both models showed comparable performance. This is due to the fact that contextual features, which are very important according to \cite{Fishcheva2021}, were not used in training XGBoost in our study.
	
	Thus we decided to take RuBERT as the classification model. After performance assessment, RuBERT with the selected optimal hyperparameters was fine-tuned on the union corpus. The obtained model was used to classify 68,859 sentences of the economic corpus.
	
	\subsection{ruGPT-3 Fine Tuning} \label{rugpt-3}
	To build a model for generating argumentative texts, fine-tuning of the pretrained model ruGPT-3 \cite{ruGPT-3} was held. To this end, 3,500 sentences (approx. 5\%) from the economic corpus annotated by the RuBERT model as “premise” with the highest confidence scores were used as training data.
	
	We fine-tuned the ruGPT3Large version (760M parameters) using NVIDIA RTX A6000 video card. The training sentences, which were prefixed with “\begin{otherlanguage}{russian} \textit{потому что}\end{otherlanguage}” (“\textit{because}”), were fed individually to the input of the model. We tuned hyperparameters using hold-out strategy: 3,000 sentences were used as training dataset, 500 sentences -- as validation dataset. The perplexity was the performance metric.
	
	The hyperparameters were as follows:
	\begin{itemize}
	    \item number of epochs = [1..5],
	    \item batch size = [1, 2, 4, 8, 12, 16],
	    \item learning rate = [$10^{-6}$, $10^{-5}$, $5\cdot10^{-5}$, $10^{-4}$].
	\end{itemize}
	
	The following hyperparameters turned out to be the best: number of epochs 1, batch size 12, learning rate $5\cdot10^{-5}$. Perplexity with these hyperparameters is equal to 9.66.
	
	At the next step, the fine-tuned ruGPT-3 model was tested by manual annotation of the generated arguments.
	
	\subsection{Manual Annotation of Generated Premises} \label{manual_annotation}
	To test the fine-tuned ruGPT-3 model, we used 20 prompting claims (after each claim, suffix “\begin{otherlanguage}{russian} \textit{потому что}\end{otherlanguage}” -- “\textit{because}” was added) (Table~\ref{tab_prompts}).The original ruGPT3Large model was used as the baseline (zero-shot learning). For each prompt (with suffix “\begin{otherlanguage}{russian} \textit{потому что}\end{otherlanguage}”), each model generated 20 sentences – potential premises. The following parameters were used for generation in both models: \textit{Top-K}=50, \textit{Top-p}=0.92. Thus, each model generated 400 sentences.
	
	\begin{table}[h!]
    	\begin{center}
        	\caption{Prompting claims}\label{tab_prompts}
        	\renewcommand\tabcolsep{5.0pt}
        	\newcolumntype{Y}{>{\raggedright\arraybackslash}X}
        	\begin{tabularx}{\textwidth}{| c | Y | Y |}
        	\hline
			No. & Prompting claims (Russian) & English translation\\
			\hline
			1 & \phantom{}\begin{otherlanguage}{russian}Банкам следует более широко использовать биометрию\end{otherlanguage}& Banks should use biometrics more widely\\
			2 & \phantom{}\begin{otherlanguage}{russian} Вместо потребительского кредита лучше взять автокредит\end{otherlanguage} & Instead of a consumer loan, it is better to take a car loan\\
			3 & \phantom{}\begin{otherlanguage}{russian} В случае высокой инфляции нужно вкладываться в драгоценные металлы\end{otherlanguage} & In case of high inflation, you need to invest in precious metals\\
			4 & \phantom{}\begin{otherlanguage}{russian} Государственные облигации являются одним из наиболее надежных видов ценных бумаг\end{otherlanguage} & Government bonds are one of the most reliable types of securities\\
			5 & \phantom{}\begin{otherlanguage}{russian} Деньги нужно вкладывать в акции\end{otherlanguage} & Money should be invested in stocks\\
			6 & \phantom{}\begin{otherlanguage}{russian} Деньги нужно вкладывать в облигации\end{otherlanguage} & Money should be invested in bonds\\
			7 & \phantom{}\begin{otherlanguage}{russian} Для сохранения финансов оптимальнее всего использовать депозиты\end{otherlanguage} & To save finances, it is best to use deposits\\
			8 & \phantom{}\begin{otherlanguage}{russian} Криптовалюты лучше фиатных валют\end{otherlanguage} & Cryptocurrencies are better than fiat currencies\\
			9 & \phantom{}\begin{otherlanguage}{russian} Лучшая инвестиционная стратегия - негосударственные пенсионные фонды\end{otherlanguage} & The best investment strategy - non-state pension funds\\
			10 & \phantom{}\begin{otherlanguage}{russian} Лучшей инвестицией является покупка недвижимости\end{otherlanguage} & The best investment is the purchase of real estate\\
			11 & \phantom{}\begin{otherlanguage}{russian} Наиболее выгодно вкладываться в голубые фишки \end{otherlanguage}& The most profitable investment in blue chips\\
			12 & \phantom{}\begin{otherlanguage}{russian} Не следует играть на валютном рынке\end{otherlanguage} & You should not play in the foreign exchange market\\
			13 & \phantom{}\begin{otherlanguage}{russian} Обучение финансовой грамотности зачастую приводит к необоснованной уверенности\end{otherlanguage} & Financial literacy training often leads to unwarranted confidence\\
			14 & \phantom{}\begin{otherlanguage}{russian} Покупка земельного участка является хорошей инвестиционной стратегией\end{otherlanguage}& Buying a plot of land is a good investment strategy\\
			15 & \phantom{}\begin{otherlanguage}{russian} При оформлении кредитной карты стоит внимательно отнестись к выбору банка\end{otherlanguage} & When applying for a credit card, you should carefully consider the choice of bank\\
			16 & \phantom{}\begin{otherlanguage}{russian} Сбережения следует хранить в валюте\end{otherlanguage} & Savings should be kept in foreign currency\\
			17 & \phantom{}\begin{otherlanguage}{russian} Сбережения следует хранить в долларах\end{otherlanguage} & Savings should be kept in dollars\\
			18 & \phantom{}\begin{otherlanguage}{russian} Сбережения следует хранить в евро\end{otherlanguage} & Savings should be kept in euros\\
			19 & \phantom{}\begin{otherlanguage}{russian} Сбережения следует хранить в рублях\end{otherlanguage} & Savings should be kept in rubles\\
			20 & \phantom{}\begin{otherlanguage}{russian} Храните деньги в той валюте в которой получаете зарплату\end{otherlanguage} & Keep money in the currency in which you receive your salary\\
			\hline
        	\end{tabularx}
		\end{center}
	\end{table}
	
	The overall generated 800 sentences were manually annotated by four annotators. Annotators were provided with <\textit{claim, sentence}> pairs with no information about which model generated the sentence. An argument was considered to be a sentence that could be used to convince an opponent of a given claim. The final decision about the sentence was based on a simple voting of the annotators' scores -- the label was approved if at least 3 from 4 annotators agreed.
	
	As a result, the labels were assigned to 660 sentences (82.5\% from 800 sentences) (Table~\ref{tab_results}): 321 for the fine-tuned model and 339 for the original model. The agreement of the annotators calculated by Krippendorf alpha=0.4772. This level of agreement corresponds to moderate agreement on the Landis \& Koch scale \cite{Landis1977}.
	
		\begin{table}[h!]
    	\begin{center}
        	\caption{Results of premise generation}\label{tab_results}
        	\renewcommand\tabcolsep{5.0pt}
        	\newcolumntype{Y}{>{\raggedright\arraybackslash}X}
        	\begin{tabularx}{\textwidth}{| X | X | X | X |}
			\hline
			Model & Premise & Non-premise & Sum \\
			\hline
			Original model & 144 / 42.5\% & 195 / 57.5\% & 339 / 100\% \\
			\hline
			Fine-tuned model & \textbf{203 / 63.2\%} & 118 / 36.8\% & 321 / 100\%  \\
			\hline
			\hline
			Overall & 347 / 52.6\% & 313 / 47.4\% & 660 / 100\% \\
			\hline
        	\end{tabularx}
		\end{center}
	\end{table}
	
	After manual annotation, the accuracy of the fine-tuned ruGPT-3 model was 63.2\% (203 premises out of 321 sentences), and the accuracy of the original model was 42.5\% (144 premises out of 339 sentences). Examples of premises generated by both models are shown in Table~\ref{tab_examples_ru} (English version is presented in Table~\ref{tab_examples_en}).
	
	\subsection{Ethical Considerations}
	\label{ethical_considerations}
	Since the proposed method is not controlled, the model may distort the meaning of the texts used in its fine-tuning. Therefore, the generated arguments may contain incorrect information and mislead users who are not experts. At the same time, large language generation models, including the RuGPT-3 model used in the proposed method, make it possible to generate texts that are indistinguishable from texts written by a person. Given the persuasive nature of the arguments generated by the proposed method, they can be used to deceive or even spread personalized disinformation \cite{Zellers2019,Schuster2020}. We hope that such technologies will be used only for peaceful purposes to improve and simplify communications in society, and that the openness and independence of the development of such models will not only ensure equal access to them, but also stimulate the development of methods for detecting the use of such models.
	
		\begin{table}[h!]
		\begin{center}
			\caption{Examples of text generation (in Russian): claims (suffix “\begin{otherlanguage}{russian}\textit{потому что}\end{otherlanguage}” is not shown); generated sentences -- premises; annotator scores’ voting results.
			}\label{tab_examples_ru}
			\renewcommand\tabcolsep{5.0pt}
			\begin{tabularx}{\textwidth}{| l | X | c |}
				\hline
				Claims & Sentences & Voting\\
				\hline
				\multirow{2}{4cm}{\begin{otherlanguage}{russian} \textit{Банкам следует более широко использовать биометрию}\end{otherlanguage}} & \underline{Fine-tuned ruGPT-3}: \newline
				\begin{otherlanguage}{russian} \textit{она позволяет упростить процедуру взаимодействия с клиентами и упрощает проведение платежа.}\end{otherlanguage}
				& 4 from 4 \\
				& \underline{Original ruGPT-3}: \newline
				\begin{otherlanguage}{russian} \textit{она может значительно упростить получение и обработку данных о клиентах.}\end{otherlanguage}
				& 4 from 4 \\ \hline
				\multirow{2}{4cm}{\begin{otherlanguage}{russian} \textit{Государственные облигации являются одним из наиболее надежных видов ценных бумаг}\end{otherlanguage}} 
				& \underline{Fine-tuned ruGPT-3}: \newline
				\begin{otherlanguage}{russian} \textit{они не подвержены обесценению, не утрачивают своей ликвидности, их стоимость остается стабильной.}\end{otherlanguage}
				& 4 from 4 \\
				& \underline{Original ruGPT-3}: \newline
				\begin{otherlanguage}{russian} \textit{они обеспечены государственной гарантией, которая включает в себя государственный долг.}\end{otherlanguage}
				& 4 from 4 \\ \hline
				\multirow{2}{4cm}{\begin{otherlanguage}{russian} \textit{Обучение финансовой грамотности зачастую приводит к необоснованной уверенности}\end{otherlanguage}} 
				& \underline{Fine-tuned ruGPT-3}: \newline
				\begin{otherlanguage}{russian} \textit{человек, желающий научиться работать с цифровыми активами, начинает думать, что он сможет легко ими управлять.}\end{otherlanguage}
				& 3 from 4 \\
				& \underline{Original ruGPT-3}: \newline
				\begin{otherlanguage}{russian} \textit{люди не в состоянии проверить, насколько достоверны и правильны их утверждения.}\end{otherlanguage}
				& 3 from 4 \\ 
				\hline
			\end{tabularx}
		\end{center}
	\end{table}
	
	\begin{table}[h!]
		\begin{center}
			\caption{Examples of text generation (English version of Table~\ref{tab_examples_ru}): claims (suffix “\textit{because}” is not shown); generated sentences -- premises; annotator scores’ voting results.
			}\label{tab_examples_en}
			\renewcommand\tabcolsep{5.0pt}
			\begin{tabularx}{\textwidth}{| l | X | c |}
				\hline
				Claims & Sentences & Voting\\
				\hline
				\multirow{2}{4cm}{\textit{Banks should use biometrics more widely}} & \underline{Fine-tuned ruGPT-3}: \newline
				\textit{it allows you to simplify the procedure of interaction with customers and simplifies the payment process.}
				& 4 from 4 \\
				& \underline{Original ruGPT-3}: \newline
				\textit{it can greatly simplify the acquisition and processing of customer data.}
				& 4 from 4 \\ \hline
				\multirow{2}{4cm}{\textit{Government bonds are one of the most reliable types of securities}} 
				& \underline{Fine-tuned ruGPT-3}: \newline
				\textit{they are not subject to depreciation, do not lose their liquidity, their value remains stable.}
				& 4 from 4 \\
				& \underline{Original ruGPT-3}: \newline
				\textit{they are backed by a state guarantee, which includes the public debt.}
				& 4 from 4 \\ \hline
				\multirow{2}{4cm}{\textit{Financial literacy training often leads to unwarranted confidence}} 
				& \underline{Fine-tuned ruGPT-3}: \newline
				\textit{a person who wants to learn how to work with digital assets begins to think that he can easily manage them.}
				& 3 from 4 \\
				& \underline{Original ruGPT-3}: \newline
				\textit{people are not able to check how reliable and correct their statements are.}
				& 3 from 4 \\ 
				\hline
			\end{tabularx}
		\end{center}
	\end{table}
	
	\section{Conclusion}
	In our study, we proposed an approach for creating a Russian-language model of premise generation for a given claim in the economic domain. First, we compared the RuBERT and XGBoost models on the translated argument corpora ArgMicro, PersEssays, and UKP Sentential. The performance of RuBERT turned out to be better. The fine-tuned RuBERT model was used to annotate sentences from a corpus of economic news. The ruGPT-3 model was fine-tuned on the annotated sentences. The model’s ability to generate argumentative premises was tested using manual annotation. The accuracy of generation of the fine-tuned model was more than 20 percentage points higher than that of the original model (63.2\% vs. 42.5\%), which confirms the effectiveness of the proposed approach. We made this corpus of 660 manually annotated pairs <\textit{claim, sentence}> publicly available\footnote{\href{https://github.com/kotelnikov-ev/economic_argument_generation}{https://github.com/kotelnikov-ev/economic\_argument\_generation}}.
	
	In our opinion, the achieved quality of argumentative text generation by the fine-tuned models allows us to speak about the possibility of using such models in practice, for example, for online generation of arguments during business meetings.
	
	As perspective areas of research, we plan to study the input context influence on GPT-type models quality.

\section*{Acknowledgements}

This work was supported by Russian Science Foundation, project № 22-21-00885, \href{https://rscf.ru/en/project/22-21-00885}{https://rscf.ru/en/project/22-21-00885}.

\bibliography{refs.bib}
\bibliographystyle{dialogue}



\end{document}